\documentclass[10pt, a4paper]{article}

\usepackage{lrec-coling2024} % this is the new style
\usepackage{tabularx}
\usepackage{multirow}
\usepackage{natbib}

\title{Evaluating Shortest Edit Script Methods for Contextual Lemmatization}

\name{Olia Toporkov, Rodrigo Agerri}

\address{HiTZ Center - Ixa, University of the Basque Country UPV/EHU \\
         \{olia.toporkov,rodrigo.agerri\}@ehu.eus\\}

\abstract{Modern contextual lemmatizers often rely on automatically induced Shortest Edit Scripts (SES), namely, the number of edit operations to transform a word form into its lemma. In fact, different methods of computing SES have been proposed as an integral component in the architecture of several state-of-the-art contextual lemmatizers currently available. However, previous work has not investigated the direct impact of SES in the final lemmatization performance. In this paper we address this issue by focusing on lemmatization as a token classification task where the only input that the model receives is the word-label pairs in context, where the labels correspond to previously induced SES. Thus, by modifying in our lemmatization system only the SES labels that the model needs to learn, we may then objectively conclude which SES representation produces the best lemmatization results. We experiment with seven languages of different morphological complexity, namely, English, Spanish, Basque, Russian, Czech, Turkish and Polish, using multilingual and language-specific pre-trained masked language encoder-only models as a backbone to build our lemmatizers. Comprehensive experimental results, both in- and out-of-domain, indicate that computing the casing and edit operations separately is beneficial overall, but much more clearly for languages with high-inflected morphology. Notably, multilingual pre-trained language models consistently outperform their language-specific counterparts in every evaluation setting.
 \\ \newline  }

\begin{document}

\maketitleabstract

\section{Introduction}

Lemmatization is one of the most common basic Natural Language Processing (NLP) tasks,
commonly understood as transforming an inflected wordform (e.g., \emph{feeling, felt})
into its initial form known as lemma (e.g., \emph{feel}), as defined by the contextual lemmatization SIGMORPHON 2019 shared task \cite{aiken-etal-2019-sigmorphon}.

Lemmatization remains important for morphologically rich languages as it usually plays a crucial role for information extraction systems, sentiment analysis and helps to deal with inflected named entities during named entity recognition task, especially for high-inflected languages.

Nowadays, state-of-the-art approaches to lemmatization are based on supervised contextual methods, a technique first proposed by \citet{chrupala-etal-2008-learning}. Treating lemmatization as a supervised classification task relies on automatically inducing a set of patterns from textual corpora, encoding the minimum amount of edits needed to map the surface word to its lemma, namely, the Shortest Edit Script (SES). Ideally these SES would capture morphological patterns about word inflection making lemmatization feasible as a classification task. Thus, in Chrupala's approach, classifiers would learn previously induced SES which, at inference time, would be decoded back into their lemmas.

Modern contextual lemmatizers often rely on automatically induced Shortest Edit Scripts (SES) for optimal performance. In fact, different methods of computing SES have been proposed as an integral component in the architecture of several state-of-the-art contextual lemmatizers currently available \cite{malaviya-etal-2019-simple,straka-etal-2019-udpipe,yildiz-tantug-2019-morpheus}. However, previous work has not investigated the direct impact of SES in the final lemmatization performance. In order to address this issue, in this paper we compare three popular approaches to automatically induce SES \citep{straka-etal-2019-udpipe,yildiz-tantug-2019-morpheus,agerri-etal-2014-ixa,agerri_rigau} and empirically investigate which of them (if any) is the most beneficial. 

In order to do so, we follow previous work by \citet{toporkov-agerri} which demonstrates that language models can competitively perform contextual lemmatization without receiving any explicit morphological signal during training, using just the word form and its corresponding SES. This allows us to focus on lemmatization as a token classification task where the only input that the model receives is the word-label pairs in context, in other words, the labels corresponding to previously induced SES. Thus, by modifying in our lemmatization systems only the SES labels that the model needs to learn, we 
may then be able to objectively conclude which SES representation helps to produce the best lemmatization results.

For our experiments we pick seven languages of different morphological complexity, namely, English, Spanish, Basque, Russian, Czech, Turkish and Polish. Moreover, we use a number of multilingual and language-specific pre-trained masked language encoder-only models as a backbone to build our lemmatizers. 
To the best of our knowledge, this is the first systematic evaluation of the impact of the
SES representations for contextual lemmatization. 

Comprehensive experimental results, both in- and out-of-domain, indicate that computing the casing and edit operations separately, as proposed by UDPipe, is the best method to obtain SES overall, particularly for the languages with more complex morphology. Chrupala's approach as implemented by
\citet{agerri-etal-2014-ixa} performs 
as a close second, while the Morpheus
method \citep{yildiz-tantug-2019-morpheus} is the less optimal one. In addition, our results show that multilingual pre-trained language models consistently outperform their language-specific counterparts in every evaluation setting. This is consistent with previous research comparing monolingual and multilingual encoder-only models \citep{Agerri2022LessonsLF}.

Furthermore, our experimental setting shows how to easily obtain
competitive lemmatization results for the languages of our choice. Finally, we offer a word on contamination of language models, concluding that the results reported in this paper are not spuriously high due to model contamination.

Code, data and fine-tuned models are publicly available to facilitate further research on this topic and reproducibility of the results.\footnote{\url{https://github.com/hitz-zentroa/ses-lemma}}

\section{Related Work}

Attempts to resolve the lemmatization task started with systems based on
dictionary lookup and/or finite set of rules
\citep{karttunen-etal-1992-two,oflazer-1993-two,eustagger-1996-alegria,Segalovich2003AFM,carreras-2004-freeLing,stroppa-yvon-2005-analogical}.
These systems, apart from being language dependent, required a lot of
effort, linguistic knowledge and manual intervention, especially for more
complex languages with a high level of inflection. The creation of large
annotated corpora, which included morpho-syntactic features and lemmas, led to
the development of machine learning approaches to lemmatization in a variety of
languages. Thus, initiatives such as the Universal Dependencies
\citep{nivre-etal-2017-universal} and the UniMorph project
\citep{mccarthy-etal-2020-unimorph} allowed to gather annotated corpora in more
than 118 languages, including low-resourced and endangered ones. 

The hypothesis that context is beneficial in the case of unseen and ambiguous
words incentivized the appearance of supervised contextual lemmatizers. One of
the pioneer works in this field is the statistical contextual lemmatizer Morfette
\citep{chrupala-etal-2008-learning}. It is based on a pipeline approach and
uses a Maximum Entropy classifier to predict morphological tags and lemmas.
Crucially, \citet{chrupala-etal-2008-learning} presents for the first time the
idea of treating lemmatization as
a classification task by predicting the Shortest Edit Script (SES), namely, the shortest
sequence of instructions (insertions, deletions or replacements) needed to
transform a reversed inflected word to its lemma. The work of
\citet{chrupala-etal-2008-learning} inspired the development of many methods
for contextual lemmatization, which most of the time included the idea of using minimum edit scripts.
Among others, the IXA pipes system \citep{agerri-etal-2014-ixa,agerri_rigau} and Lemming
\citep{muller-etal-2015-joint} apply the same principle of edit trees,
combining it with the possibility of adding external lexical information. Other
examples of the systems that use the concept of SES are the works of
\citet{gesmundo-samardzic-2012-lemmatisation},
\citet{chakrabarty-etal-2017-context} and the system of
\citet{malaviya-etal-2019-simple}.

The development of supervised approaches involving deep learning algorithms and
the appearance of the Transformer architecture \citep{Vaswani2017AttentionIA} and
Transformer-based masked language models (MLMs) such as BERT
\citep{devlin-etal-2019-bert} and XLM-RoBERTa
\citep{conneau-etal-2020-unsupervised} allowed to significantly improve the
performance of supervised lemmatizers. Thus, in the SIGMORPHON 2019 shared
task on contextual lemmatization \citep{mccarthy-etal-2019-sigmorphon}
most of the participating systems were based on MLMs. The best overall
system was UDPipe
\citep{straka-etal-2019-udpipe}, which ensembled various pre-trained contextualized BERT
and Flair embeddings as an additional input to a Bi-LSTM network. To perform
lemmatization they classify the input words according to the set of generated
lemma rules or SES. The third best model, Morpheus, proposed a two-level LSTM network 
\citep{yildiz-tantug-2019-morpheus} which used vector-based representations of
words, morphological tags and SES as input. The output of the system results in
a corresponding morphological labels and SES representing the lemma class which
is later decoded into its lemma form.

However, while many of these top performing systems included different methods to compute SES as an integral component in their lemmatization models, there has not been an attempt to compare and establish which of the existing methods is the optimal one for the task. In this paper we pick three of the most popular SES approaches (according to performance on the SIGMORPHON 2019 lemmatization benchmark) and make a systematic comparison with the aim of clarifying this issue. We believe that this could benefit the development of future lemmatizers which may include SES as an integral component of their systems.

\section{Data}

To train and evaluate our models we used the datasets developed for the
SIGMORPHON 2019 shared task on contextual lemmatization
\citep{mccarthy-etal-2019-sigmorphon}. These datasets are annotated according
to the Unimorph schema guidelines \citep{mccarthy-etal-2020-unimorph}. For in-domain evaluation we chose one corpus per language with standard train and development partitions. Additionally, we also provide out-of-domain evaluation results, as this is the setting in which lemmatizers are usually deployed. As most languages are represented in the SIGMORPHON 2019 by more than one dataset, for out-of-domain evaluation we picked the test sets of datasets different from the ones selected for in-domain evaluation. The exception was Basque, for which we selected a dataset external to the UniMorph SIGMORPHON data.

With respect to in-domain, in the case of Spanish and Russian we used GSD data, which consists of Wikipedia and news articles, texts from blogs and reviews. As the lemmas of these two corpora were originally lower-cased and giving the fact that the methods of generating the Shortest Edit Scripts (SES) are case dependent,
we performed a simple adjustment by changing the lemmas of the proper nouns to
their upper-cased version. For the rest of the languages, there was no need of performing such adjustment, as the lemmas for the proper nouns in the corresponding corpora were correctly upper-cased by default. 

For English we chose English Web Treebank (EWT) \citep{silveira-etal-2014-gold} composed using different Web sources, such as several media, blog articles, reviews, e-mails and Yahoo! answers.

For Basque we used the Basque Dependency Treebank (BDT) \citep{bdt_corpus} made of literary and journalistic texts.

With respect to Turkish we used ITU-METU-Sabanci Treebank (IMST) \citep{turk-etal-2019-turkish}, a corpus formed by sentences from a wide range of domains, such as non-fiction and news.

The Czech data correspond to the CAC corpus \citep{czech_cac_corpus}, based on the Czech Academic Corpus 2.0, containing mostly full-length articles from different media sources, such as newspapers, magazines and transcripts of spoken language from radio and TV programs.

Finally, for Polish we chose the LFG corpus \citep{przepiorkowski-patejuk-2018-lexical}, derived from a corpus of LFG (Lexical Functional Grammar) syntactic structures, and consisting mostly of sentences from fiction, news and non-fiction genres, as well as the texts from the Internet sources. 

Out-of-domain evaluation was performed using the AnCora corpus \citep{taule-etal-2008-ancora} in the case of Spanish, which consists mainly of the news texts. For Russian we chose SynTagRus \citep{syntagrus}, a corpus that is formed using texts from popular science, news and journal articles and contemporary fiction. Regarding English we used the Georgetown University Multilayer (GUM) corpus \citep{Zeldes2017}, a collection of a wide range of Web texts from Wikipedia, Reddit and Wikinet. As for the Basque language, due to unavailability of the alternative corpora provided in the shared task data, we chose the Armiarma corpus \citeplanguageresource{armiarma_corpus}, which was created using literary reviews. 
To evaluate Turkish and Czech languages we used PUD data, a part of the Parallel Universal Dependencies treebanks created for the CoNLL
2017 shared task \citep{zeman-etal-2017-conll}. The corpora include sentences from such domains as Wikipedia and news, annotated in total for 18 languages. Table \ref{tab:data_details} provides details about the size of the datasets.

\begin{table}[]
\begin{tabular}{l|r|r|r|r}
\hline
   & \multicolumn{1}{c|}{train} & \multicolumn{1}{c|}{dev} & \multicolumn{1}{c}{test} & \multicolumn{1}{|c}{test(OOD)} \\ \hline
es & 345,545 & 42,545 & 43,497 & 54,449  \\
ru & 79,989 & 9,526  & 9,874  & 109,855  \\
en & 204,857 & 24,470 & 25,527 & 8,189   \\
eu & 97,336  & 12,206 & 11,901 & 299,206 \\
tr & 46,417  & 5,708  & 5,734  & 1,795   \\
cz & 395,043 & 50,087 & 49,253 & 1,930   \\
pl & 104,730 & 13,161 & 13,076 & 8,511  \\
\hline
\end{tabular}
 \caption {Number of tokens in the train, development, in-domain (test) and out-of-domain (test(OOD)) test sets.}
 \label{tab:data_details}
\end{table}

\section{Methods to Induce Shortest Edit Scripts} \label{sec:ses_types}

The general idea of computing the Shortest Edit Script (SES) in contextual lemmatization is based on
finding the minimum number of edit operations necessary to convert a surface
word into its corresponding lemma. By edit operations we understand any change applied to the wordform, which consists in deleting, inserting or replacing letters in the surface word, as well as leaving the word unchanged in the case the inflected form and the lemma coincide (e.g. the$\rightarrow$\emph{the}, road$\rightarrow$\emph{road}). SES methods focus on codifying such minimum edits for their further application as a set of instructions to modify the surface word. In this paper we address three different approaches based on the Shortest Edit Scripts. The methods chosen are those implemented by the first and
third best systems in the SIGMORPHON 2019 shared task, namely UDPipe
\citep{straka-etal-2019-udpipe} and Morpheus 
\citep{yildiz-tantug-2019-morpheus}, and Chrupala's original proposal as
implemented by the IXA pipes system \citep{agerri-etal-2014-ixa,agerri_rigau}. \footnote{It may be argued that the methods of Morpheus and UDPipe systems do not strictly always generate the \emph{shortest} edit script (SES). However, we keep the SES term as it was originally coined by \citet{chrupala-etal-2008-learning} as a convenient acronym.}

\subsection{UDPipe}

The approach applied in the UDPipe system focuses on performing character level
edits on the suffixes and prefixes of the word. They divide their script
creation in two parts: (i) encoding the correct casing as a casing script and
(ii) creating a sequence of character edits. Regarding the casing script,
they consider both wordform and lemma as lower-cased. If the lemma contains
upper-cased characters they add a rule to the casing script to uppercase such
characters in the final lemma. The next step is creating a sequence of
character edits by splitting the wordform into a prefix, a root (stem) and a
suffix in order to process them separately. The root is obtained by finding the
longest equal substring between the input word and its lemma and is kept
unchanged. Then they process the prefixes and suffixes of the target word, including
possible character operations such as \emph{copy}, \emph{add} or \emph{delete}.
The final script is produced by a concatenation of the casing and the edit
scripts. The obtained SES are the complete rules which convert input words to
their lemmas. When the word and lemma do not share any common parts, the
word is considered irregular and is directly replaced by its lemma, skipping
any possible edits. 

\subsection{Morpheus}

Morpheus's approach is based on the prediction of minimum edits 
between a surface word and its lemma using four fundamental operations
such as \emph{same}, \emph{delete}, \emph{replace} and \emph{insert}.
\emph{Same} and  \emph{delete} operations have only one version (the character
may be left without changes (s) or deleted (d)), while \emph{replace} and
\emph{insert} operations may vary, depending on the character they are tied to.
As the minimum edit prediction decoder of Morpheus creates edit labels for each
character in the word, it is only able to generate lemmas shorter or equal to
the inflected forms. Still, in some languages lemmas may be longer that their
corresponding wordforms. For such cases \citet{yildiz-tantug-2019-morpheus} modify the standard
Levenshtein distance algorithm by merging successive \emph{insert} labels into
one in the same position with multiple characters. They perform the same
process for the \emph{replace} label, combining it with the successive
\emph{insert} labels into one \emph{replace} label and ensuring the correct
lemma generation. They also consider the cases where the word is situated in the beginning of the sentence and should be lowercased, reflecting it in the Shortest Edit Script.

\subsection{IXA pipes}

The third method is based on the interpretation of Chrupala's technique
\citep{chrupala-etal-2008-learning} by \citet{agerri-etal-2014-ixa}. 
This approach addresses the suffixal
nature of inflectional morphology where the end of the word is the most
changing part and is more likely subject to modifications than the prefix or root.
\citet{chrupala-etal-2008-learning} propose to compute the minimum edit distance
between the \emph{reversed} wordform and its lemma. They index word characters
starting from the end of the string, allowing to form more coherent lemma
classes and to perform lemmatization more efficiently. In the set of
instructions that are generated using this technique the position of the letters that are subject to change are indicated along with the type of operation (insertion or deletion). In the adaptation of this approach it is also considered the casing of proper nouns, as well as the casing of the words that appear in the beginning of the sentence and should be lowercased for their correct lemmatization.

\subsection{SES Comparison}

In order to obtain a better understanding of the described methods and their core differences, we provide a brief comparison of the three minimum edit approaches, namely, UDPipe system's approach (\emph{ses-udpipe}), Morpheus's approach (\emph{ses-morpheus}) and IXA pipes approach (\emph{ses-ixapipes}).  

\begin{table}[h]
\footnotesize
     \centering
\begin{tabular}{l | l | l | l }
 \hline
\textbf{word$\rightarrow$\emph{lemma}} & \textbf{UDP} & \textbf{IXA} & \textbf{Morph} \\ \hline 
cats$\rightarrow$cat & ↓0;d¦- & D0s & s|s|s|d \\
birds$\rightarrow$\emph{bird} & ↓0;d¦- & D0s & s|s|s|s|d \\
did$\rightarrow$\emph{do} & ↓0;d¦--+o & R1ioD0d & s|r\_o|d \\
Wolak$\rightarrow$\emph{Wolak} & ↑0¦↓1;d¦ & O & s|s|s|s|s \\
You$\rightarrow$\emph{you} & ↓0;d¦ & 1 & l|s|s \\
\hline
\end{tabular}
    \caption {Examples of the three types of SES patterns: UDP - ses-udpipe, IXA - ses-ixapipes, Morph - ses-morpheus.}
    \label{tab:ses_examples}
\end{table}

Table \ref{tab:ses_examples} provides some examples of the Shortest Edit Scripts used in lemmatization for the aforementioned SES methods. For an action such as removing the last letter of the surface word (as in the case of the words `cats' and `birds' ) both \emph{ses-udpipe} and \emph{ses-ixapipes} apply the edit instruction to the reversed wordform, removing the last letter. Additionally, \emph{ses-udpipe} method indicates that the word has to be lowercased. As for \emph{ses-morpheus}, it processes each letter separately, leaving those that should remain untouched as `s' (same) and deleting the last one, marking such operation with `d' (delete). Unlike \emph{ses-udpipe} and \emph{ses-ixapipes}, the scripts corresponding to the same action of deleting the last word's letter generate two different label classes as the number of the letters in `cats' and `birds' is distinct. 

The next lemmatization example (did$\rightarrow$\emph{do}), demonstrates how each of the SES approaches treats the cases where one or more letters should be inserted in order to obtain the lemma. Here \emph{ses-ixapipes} and \emph{ses-morpheus} methods apply similar order of minimum edits using delete and replace operations, while \emph{ses-udpipe} first deletes the two ultimate letters of the word and only then makes the insertion of the letter `o'. 

Finally, the last two examples are provided in order to reflect the edit scripts that are generated in the case of proper nouns in contrast to the ordinary nouns situated in the beginning of the sentence and, therefore, starting with the capital letter. We could see that for the proper noun `Wolak' \emph{ses-udpipe} indicates that the first letter should remain uppercased, whereas the scripts of \emph{ses-ixapipes} and \emph{ses-morpheus} simply leave the word unchanged. As for the pronoun `You' situated in the beginning of the sentence, all three SES approaches lowercase it in order to obtain the correct lemma. It is important to mention that, as with the first two examples, in the case of the longer proper nouns the UDPipe's and IXA pipes' scripts would remain the same, while the script of the Morpheus's approach would vary according to the number of letters in the surface word.

\section{Systems}

In our experiments we apply two multilingual and seven language-specific
pre-trained masked language models (MLMs). With respect to multilingual models we use multilingual BERT
(mBERT) \citep{devlin-etal-2019-bert}, a Transformer-based masked language
model pre-trained on the Wikipedias of 104 languages. mBERT was pre-trained using
both masking and next sentence prediction objectives, applying a batch size of 256 and
512 sequence length for 1M steps. The
second multilingual model we apply is XLM-RoBERTa 
\citep{conneau-etal-2020-unsupervised}, pre-trained on 2.5TB of filtered
CommonCrawl data for 100 languages. This model is based on the RoBERTa
architecture, was trained only on the MLM task, implies dynamic
mask generation and was pre-trained over 1.5M steps with a batch of 8192 and sequences
of 512 length. We used both base and large versions of XLM-RoBERTa.

Regarding the language-specific models, we choose one model for each of the target languages. For Spanish we use the cased version of BETO \citep{CaneteCFP2020}. It is a BERT-base language model trained on a large Spanish corpus including all Spanish Wikipedia as well as the Spanish part of the OPUS project \citep{tiedemann-2012-parallel} in a total size of around 3 billion words. BETO is an upgraded version of the initial BERT-base model with the application of the dynamic masking technique, introduced in RoBERTa. It was trained with the total of 2M steps in two stages: 900K steps with a batch size of 2048 and maximum sequence length of 128, and the rest of the steps using batch size of 256 and maximum sequence length of 512.

For the Czech language we apply slavicBERT \citep{arkhipov-etal-2019-tuning},  developed by continuing the training of multilingual BERT on Russian news and the Wikipedias of Russian, Bulgarian, Czech and Polish. The vocabulary of subword tokens was also rebuilt with the use of the subword-nmt repository.\footnote{\url{https://github.com/rsennrich/subword-nmt/}} 

For Russian we choose RuBERT \citep{rubert-kuratov}, which was developed similarly to slavicBERT, with the difference of having only Russian as the target language. The system was trained using the Russian Wikipedia and news. The authors obtain a new subword vocabulary with longer Russian words and subwords from subword-nmt. 

In the case of English we train RoBERTa-base \citep{liu2019roberta}, an optimized version of the BERT model. This model was obtained using more than 160GB of uncompressed text, including, apart from the standard BERT datasets, the CC-news dataset with English news articles published between January 2017
and December 2019.

For the Polish language we apply the base version of HerBERT \citep{mroczkowski-etal-2021-herbert}. This model is based on the original BERT architecture and achieves state-of-the-art results on several downstream
tasks, obtaining the best overall scores for Polish language understanding on the KLEJ Benchmark. HerBERT was trained on two datasets merged from six corpora such as NKJP, Wikipedia, Wolne Lektury, CCNet and Open Subtitles. Its base version outperformed the base version of Polish RoBERTa despite being trained with a smaller batch size (2560 vs 8000) and for a fewer number of steps
(50k vs 125k).

In the case of Turkish we use BERTurk.\footnote{\url{https://github.com/stefan-it/turkish-bert}} It is a cased BERT-base model, trained on 35GB of data, including Wikipedia, various OPUS corpora \citep{tiedemann-2016-opus}, data provided by Kemal Oflazer and the version of the Turkish OSCAR corpus \citep{OrtizSuarezSagotRomary2019} which was previously filtered and sentence segmented.

Finally, for Basque we use BERTeus \citep{agerri-etal-2020-give}, a BERT-base model trained on the BMC Basque corpus, which consists of news articles from online newspapers and the Basque Wikipedia.  The authors also perform the subword
tokenization, which is closer to linguistically interpretable strings in Basque. BERTeus outperforms mBERT and XLM-RoBERTa in several NLP tasks including named entity recognition, POS tagging, sentiment analysis and topic modelling.

\section{Experimental Setup}

In order to compare the three different approaches to generate the Shortest Edit Scripts (SES) described in Section \ref{sec:ses_types}, we fine-tuned the multilingual and language-specific pre-trained masked language models for each
language in a token classification task, where the labels to be predicted correspond to the automatically induced SES. The MLMs were fine-tuned by adding a single linear classification layer on top. We performed a grid search of hyperparameters to select the best batch size (8, 16), weight decay (0.01, 0.1), learning rate (2e-5, 3e-5, 5e-5) and epochs (5, 10, 15, 20). We conduct both in-domain and out-of-domain evaluation of the models. By out-of-domain evaluation we understand evaluating on a data distribution different from the one that was used for training (in the in-domain setting). For each type of SES we chose the best model on the development set among the four MLMs in terms of word accuracy and loss. For all the languages the highest accuracy was achieved using XLM-RoBERTa large model, being the only exception the \emph{ses-morpheus} method in the case of Russian, where the best accuracy was achieved using mBERT. Thus, every result reported in the next subsections is obtained using XLM-RoBERTa-large as a backbone. Finally, apart from calculating word and sentence accuracy scores, we also report the statistical significance across the three SES methods using the McNemar test \citep{dietterich}.

\subsection{Results}

Table \ref{tab:experimental_results_wacc} reports the best overall word accuracy results for in-domain and out-of-domain settings. 
We could see that among the three SES types \emph{ses-morpheus} is the least optimal. Since its functioning principle implies that the same edit operation may generate various labels depending on the word's total number of characters (as demonstrated in Table \ref{tab:ses_examples} with the examples of the words `cats' and `birds'), this approach creates the highest amount of unique labels for 5 out of 7 languages of our survey (as illustrated by Table \ref{tab:unique_labels_SES}). This might be one of the possible reasons that leads to the lower performance of this SES method, as in this case the range of the SES classes is wider, which could difficult the learning and generalization processes of the model.

\begin{table}[h]
\centering
\footnotesize
\begin{tabular}{l|r|r|r}
\hline
   & \multicolumn{1}{l|}{ses-udpipe} & \multicolumn{1}{l|}{ses-ixapipes} & \multicolumn{1}{l}{ses-morpheus} \\ \hline
es & 444 & 670 & 1,213  \\
ru & 1,157 & 2,390 & 3,208 \\
en & 286 & 445 & 891 \\
eu & 2,247 & 5,324 & 3,710 \\
tr & 236 & 4,147 & 799 \\
cz & 1,020 & 2,345 & 3,033 \\
pl & 947 & 1,920  & 2,692 \\                                
\hline
\end{tabular}
    \caption{The amount of unique labels for each SES type.}
    \label{tab:unique_labels_SES}
\end{table}

\begin{table}[h]
\renewcommand{\footnotesize}{\fontsize{8pt}{10pt}\selectfont}
\footnotesize
\centering
\begin{tabular}{l|ll|ll|ll}
\hline
   & \multicolumn{2}{c|}{ses-udpipe} & \multicolumn{2}{c|}{ses-ixapipes} & \multicolumn{2}{c}{ses-morpheus}\\ \hline
   & \multicolumn{1}{c}{IND} & \multicolumn{1}{c|}{OOD} & \multicolumn{1}{c}{IND} & \multicolumn{1}{c|}{OOD} & \multicolumn{1}{c}{IND} & \multicolumn{1}{c}{OOD} \\ \hline
es & 0.983 & 0.971 & \textbf{0.983} & \textbf{0.972}*  & 0.975 & 0.963 \\
ru & \textbf{0.973} & \textbf{0.945}* & 0.970 & 0.941 & 0.927 & 0.885 \\
en & 0.991 & 0.939  & \textbf{0.991} & \textbf{0.941}  & 0.979 & 0.916 \\
eu & \textbf{0.969}* & \textbf{0.890}*  & 0.966 & 0.885 & 0.952 & 0.857 \\
tr & \textbf{0.964}* & \textbf{0.853}* & 0.915 & 0.827 & 0.938 & 0.804 \\
cz & \textbf{0.994}* & 0.947 & 0.991 & \textbf{0.951}  & 0.987 & 0.924 \\
pl & \textbf{0.982}* & \textbf{0.952} & 0.980  & 0.950 & 0.943 & 0.917 \\
\hline
\end{tabular}
\caption {Word accuracy results for the 3 SES types for in-domain (IND) and out-of-domain (OOD) settings. In \textbf{bold}: best overall results across systems and SES types. *:results, that are statistically significant at $\alpha = .05$.}
\label{tab:experimental_results_wacc}
\end{table}

\begin{table}[h]
\renewcommand{\footnotesize}{\fontsize{8pt}{10pt}\selectfont}
\footnotesize
\centering
\begin{tabular}{l|ll|ll|ll}
\hline
& \multicolumn{2}{c|}{ses-udpipe}  & \multicolumn{2}{c|}{ses-ixapipes} & \multicolumn{2}{c}{ses-morpheus} \\ \hline
& \multicolumn{1}{c}{IND} & \multicolumn{1}{c|}{OOD} & \multicolumn{1}{c}{IND} & \multicolumn{1}{c|}{OOD} & \multicolumn{1}{c}{IND} & \multicolumn{1}{c}{OOD} \\ \hline
es & 0.703 & 0.489 & \textbf{0.708} & \textbf{0.505}* & 0.582 & 0.397 \\
ru & \textbf{0.614} & \textbf{0.426}* & 0.604 & 0.401 & 0.314 & 0.187 \\
en & \textbf{0.890} & 0.425 & 0.888 & \textbf{0.439} & 0.773 & 0.305 \\
eu & \textbf{0.684} & \textbf{0.203}* & 0.663 & 0.195 & 0.551 & 0.150 \\
tr & \textbf{0.707}* & \textbf{0.080}* & 0.496 & 0.010  & 0.583 & 0.050 \\
cz & \textbf{0.896}* & 0.430 & 0.855 & \textbf{0.500} & 0.796 & 0.320 \\
pl & \textbf{0.876}* & \textbf{0.656} & 0.861 & 0.657 & 0.675 & 0.519 \\                 
\hline
\end{tabular}
    \caption{Sentence accuracy results for the 3 SES types for in-domain (IND) and out-of-domain (OOD) settings. In \textbf{bold}: best overall results across systems and SES types. *:results, that are statistically significant at $\alpha = .05$.}
    \label{tab:experimental_results_sentacc}
\end{table}

With respect to the other two methods, we could observe that for 5 out of 7 languages (namely, for Russian, Basque, Turkish, Czech and Polish) the highest word accuracy in-domain is achieved using \emph{ses-udpipe} approach (4 out of 5 of these results are statistically significant). However, in the case of Spanish and English the results are almost identical for both \emph{ses-udpipe} and \emph{ses-ixapipes} methods. Regarding out-of-domain, in 4 out of 7 cases \emph{ses-udpipe} is the optimal choice as well (3 statistically significant), while \emph{ses-ixapipes} benefits the Czech language and performs similar to the UDPipe's method for English and Spanish.

Still, the differences in word accuracy results for \emph{ses-udpipe} and \emph{ses-ixapipes} are very small, which makes it difficult to
distinguish between approaches. In order to obtain a clearer picture in the methods' performance we decided to additionally
compute the sentence accuracy metric as proposed for POS tagging by \citet{Manning2011PartofSpeechTF}.

As demonstrated in Table \ref{tab:experimental_results_sentacc}, sentence accuracy allows us to better distinguish between the models’ performance. First, it confirms the results regarding \emph{ses-morpheus} approach, achieving much lower accuracy for all the languages. Second, the almost equivalent results in word accuracy for English and Spanish using both \emph{ses-udpipe} and \emph{ses-ixapipes} methods are now noticeably different when evaluated using sentence accuracy. While in the case of Spanish the approach of IXA pipes seems to be more beneficial both in-and out-of-domain, for English it allows to achieve 1.4 points better in sentence accuracy out-of-domain. The same phenomenon can be observed in the case of the Czech language, with 7 points better in sentence accuracy out-of-domain for \emph{ses-ixapipes} method with respect to \emph{ses-udpipe}. The results for the rest of the languages follow the tendency obtained with the word accuracy metric, where the \emph{ses-udpipe} method scores the highest.

Although sentence accuracy results provide a clearer picture, we would like to establish whether the differences are in fact statistically significant. Thus, we perform the McNemar test to determine whether the scores obtained by \emph{ses-udpipe} and \emph{ses-ixapipes} are statistically significant or not (null hypothesis).
When evaluating word accuracy the test shows that the differences in performance of the two SES approaches mentioned above are statistically significant ($\alpha = .05$) in \emph{ses-udpipe} favor for the agglutinative languages (Basque and Turkish) both in-domain (with p-value $< 0.02$ for Basque and p-value $< 0.001$ for Turkish) and out-of-domain (with p-value $< 0.001$ for both Basque and Turkish); for Czech and Polish languages in-domain (p-value $< 0.001$ for Czech and p-value $< 0.005$ for Polish) and for Russian out-of-domain (p-value $< 0.001$). Such small p-value results indicate that the differences in performance of the models trained with different minimum edit approaches is noticeable. The test results also suggest that in the case of lemmatizing using \emph{ses-ixapipes} method the model commits a larger percentage of the errors respect to \emph{ses-udpipe}. As for \emph{ses-ixapipes}, the results are statistically significant only for Spanish in the out-of-domain setting (p-value $< 0.002$). For sentence accuracy the McNemar test results reflect the same tendency as for word accuracy. Therefore, the McNemar test indicates that \emph{ses-udpipe} approach is more beneficial in the generation of the Shortest Edit Scripts that the other two methods, at least in the proposed spectrum of languages.

\section{Discussion}

In order to make the comparison of the three Shortest Edit Script methods more complete we discuss the following points. First, we analyze the performance of the pre-trained masked language models on in-vocabulary and out-of vocabulary words. The aim of such analysis is to understand which SES approach contributes better to the generalization capabilities of the MLMs. Second, we conduct a brief error analysis in order to understand what makes UDPipe's method more successful that its two other counterparts.  Finally, we discuss model contamination issues.

\paragraph{Generalization on out-of-vocabulary words} Pre-trained masked language models, in particular XLM-RoBERTa, demonstrate good generalization abilities and are capable of achieving competitive results lemmatizing the words they did not see during the training process \citep{toporkov-agerri}. In order to check which SES approach benefits such capabilities more, we calculate word accuracy on in-vocabulary and out-of-vocabulary words, comparing how the model performs on the words it has seen during the training respect to the words it sees for the first time. Table \ref{tab:inv_oov} reports the results. 

Interestingly, all three SES approaches perform equally well on in-vocabulary words in-domain and obtain very similar results out-of-domain. Things start changing when we analyze the out-of-vocabulary performance. We can see the significant drop in the generalization capability of the models using \emph{ses-morpheus} approach, which confirms the word and sentence accuracy results. We also could see that for Spanish, English and Czech the results are better using \emph{ses-ixapipes} method, the point that reinforces the sentence accuracy results. There is also a strong correlation between the results where the differences between \emph{ses-udpipe} and \emph{ses-ixapipes} are statistically significant and how these approaches perform on unseen words. 

In any case, from an overall perspective \emph{ses-udpipe} demonstrates stronger performance, achieving the highest accuracy in-domain for 5 out of 7 languages and out-of-domain for 4 out of 7 languages both for in-vocabulary and out-of-vocabulary words. Table \ref{tab:oov_detailed} in Appendix A provides more detailed results on out-of-vocabulary statistics with respect to lemmas and SES. Thus, the overall better performance of \emph{ses-udpipe} is reinforced by having the lowest percentage rate of SES that have not been seen during training. This data indicates that the \emph{ses-udpipe} approach has better generalization capabilities.  

\begin{table}[h]
\centering
\footnotesize
\setlength{\tabcolsep}{0.3em}
\begin{tabular}{l|l|ll|ll|ll}
\hline
& & \multicolumn{2}{c|}{ses-udpipe} & \multicolumn{2}{c|}{ses-ixapipes} & \multicolumn{2}{c}{ses-morpheus} \\ \hline
& & INV  & OOV & INV  & OOV  & INV  & OOV \\ \hline
\multirow{2}*{es} & ind & 0.989 & 0.906 & \textbf{0.989} & \textbf{0.912} & 0.989 & 0.816 \\
& ood & 0.976 & 0.904 & \textbf{0.977} & \textbf{0.917}* & 0.975 & 0.807 \\ \hline
\multirow{2}*{ru} & ind & \textbf{0.995} & \textbf{0.908}  & 0.994 & 0.900 & 0.991 & 0.741 \\
& ood & \textbf{0.972} & \textbf{0.878}* & 0.972 & 0.865 & 0.967 & 0.686 \\ \hline
\multirow{2}*{en} & ind & \textbf{0.995} & \textbf{0.931} & 0.994 & 0.927 & 0.993 & 0.751 \\
& ood & \textbf{0.954} & 0.833 & 0.953 & \textbf{0.849}  & 0.954 & 0.631 \\ \hline
\multirow{2}*{eu} & ind & \textbf{0.990} & \textbf{0.852}* & \textbf{0.990} & 0.832 & 0.989 & 0.748 \\
& ood & \textbf{0.926} & \textbf{0.777}* & \textbf{0.926} & 0.757 & 0.926 & 0.645  \\ \hline
\multirow{2}*{tr} & ind & 0.991 & \textbf{0.882}* & 0.991 & 0.685 & \textbf{0.992} & 0.775 \\
& ood & \textbf{0.946}  & \textbf{0.693}* & 0.945 & 0.625 & 0.944 & 0.564 \\ \hline
\multirow{2}*{cz} & ind & \textbf{0.998}  &\textbf{0.955}* & \textbf{0.998} & 0.923 & \textbf{0.998} & 0.876 \\
& ood & 0.987 & 0.807 & \textbf{0.988} & \textbf{0.821} & 0.987 & 0.703 \\ \hline
\multirow{2}*{pl} & ind & \textbf{0.998} & \textbf{0.919}* & 0.997 & 0.909 & 0.992 & 0.742 \\
& ood & \textbf{0.981} & \textbf{0.816} & 0.981 & 0.808 & 0.974 & 0.650 \\ \hline        
\end{tabular}
    \caption{Word accuracy for in-vocabulary (INV) and out-of-vocabulary (OOV) words for in-domain (ind) and out-of-domain (ood) results. In \textbf{bold}: best results per SES and per language; *:results, that are statistically significant at $\alpha = .05$.}
    \label{tab:inv_oov}
\end{table}

In conclusion, the results of our experiments show that the \emph{ses-udpipe} method is more beneficial for the lemmatization task, especially in the case of the languages with more complex morphology. To analyze what makes this method better than its close counterpart \emph{ses-ixapipes}, we conduct a brief error analysis in an attempt to identify the most important factors that may influence performance.

\paragraph{Error Analysis}

The first noticeable advantage that is perceived in the structure of the \emph{ses-udpipe} patterns is the absence of indexing. While \emph{ses-ixapipes} misplaces some indexes, wrongly annotating them to the letters that should be deleted or replaced, \emph{ses-udpipe} approach simplifies this process by only indicating the positions of the letters that should be modified without having to map it with the corresponding index. Such misplacements usually affect the complex words that need a lot of edit operations in order to be lemmatized. 

Another important difference is how to deal with non-Latin alphabet and some language-specific letters. In the cases of such languages as Russian and Turkish these letters may cause a certain confusion during minimum edit generations as it happens to \emph{ses-ixapipes}, which sometimes assigns to the final SES pattern the letters that do not appear neither in the surface word, nor in the lemma. 

The third interesting observation is encountered mostly in the lemmatization of agglutinative languages (Basque and Turkish) and has to do with their suffixal nature. Whereas the \emph{ses-udpipe} method processes the parts of the words separately, \emph{ses-ixapipes} does not take into account this issue. Thus, \emph{ses-ixapipes} focuses on indexing the correct letters without considering if its the part of the suffix or of the root. As a result, this approach may create an alternative minimum edit script, which may map correctly, but that does not coincide with the gold standard SES. For example, when lemmatizing the Basque word \emph{folklorearen} (`of folklore', lemma \emph{folklore}), the gold standard SES would be D5rD4eD3aD0n, while in one of the predictions \emph{ses-ixapipes} offered an alternative version of SES, which is D4eD3aD2rD0n. Applying both sets of edits will deliver the same result, but as the goal of the classification task is to correctly assign the corresponding SES to its surface word, such cases are considered incorrect. 
In order to check whether this could be crucial in evaluating the overall SES performance, we calculate the total number of occurrences where the SES distinct from the gold standard delivers the correct lemma for the Basque language. Our results show that for \emph{ses-udpipe} approach there are 9 out of 11901 cases where an alternative SES leads to the same lemma (in-domain), while in the case of \emph{ses-ixapipes} the number of such occurrences is 17 out of 11901 respectively. This data indicates, that although such cases could appear, their influence on the overall result is insignificant. 

Finally, it also seems beneficial to encode the casing script as implemented in the \emph{ses-udpipe} method and which is only partially implemented in both \emph{ses-ixapipes} and \emph{ses-morpheus} approaches.

Regarding the other two minimum edit approaches, namely, \emph{ses-ixapipes} and \emph{ses-morpheus}, a brief error inspection shows that in the case of \emph{ses-ixapipes} most of the errors are of suffixal and root nature, more precisely, in the incorrect indexing or letter misplacement. Furthermore, the performance of \emph{ses-morpheus} is mainly affected by the large number of generated SES classes, which makes the classification task much more difficult. The cases where lemma is longer than wordform, and, therefore the edit operations are applied jointly, constitute between 5 and 15 of the total error rate across the inspected languages, and is another source of possible low performance of this method with respect to the other two.

\paragraph{A word on model contamination} We would like to finish by offering a word on model contamination. More specifically, we would like to discuss whether the performance of a MLM such as XLM-RoBERTa has been spuriously high because the model already saw the datasets we are experimenting with during pre-training, namely, whether XLM-RoBERTa has been
contaminated. In order to address this, we would like to clarify that CC-100, the corpus used to train XLM-RoBERTa, was constructed with CommonCrawl snapshots from between January and
December 2018. Moreover, the SIGMORPHON data was released in
2019\footnote{First GitHub commit December 19, 2018.} with the test data
including gold standard lemma and UniMorph annotations not being released until April 2019.
Finally, and most importantly, XLM-RoBERTa does not see the lemmas themselves
during training or inference, but the SES classes we automatically generate in
an ad-hoc manner for the experimentation. The datasets containing both the
words and the SES classes used have not been yet made publicly available.
Therefore, we can conclude that XLM-RoBERTa seems to generalize over
unseen words and that its performance is not justified by any form of language
model contamination.

\section{Conclusion}

In this paper, we present the first detailed systematic comparison of three popular methods to compute Shortest Edit Scripts (SES), widely used in modern contextual lemmatization models. After a comprehensive battery of experiments with various evaluation metrics and statistical tests, results indicate that \emph{ses-udpipe} 
is the optimal method for contextual lemmatization among the Shortest Edit Script approaches. Its main advantages consist in: (i) computing casing and edit operations separately; (ii) processing the wordform by morphemes and the absence of indexing, which allows to avoid the cases where there are the same letters in the suffix and the root (especially for agglutinative languages such as Basque and Turkish) and to create less ambiguous SES; (iii) better generalization capabilities, that result in obtaining less out-of-vocabulary SES and creating fewer SES labels, which benefits the models by having to learn a smaller amount of SES classes. Furthermore, our results indicate the following: (i) more metrics should be implemented in the analysis of the MLMs performance along with the word accuracy; (ii) out-of-domain evaluation should be considered as an important step as it allows to obtain a clearer picture of how far the task is solved.

We believe that the results of our study could be useful for the future development of contextual lemmatizers which may include SES as an integral component of their systems.

\section{Acknowledgements}

This work has been partially supported by the HiTZ center and 
the Basque Government (Research group funding IT-1805-22). We are also thankful to several MCIN/AEI/10.13039/501100011033 projects: (i) DeepKnowledge (PID2021-127777OB-C21) DeepKnowledge (PID2021-127777OB-C21) and by FEDER, EU; (ii) Disargue (TED2021-130810B-C21) and European Union NextGenerationEU/PRTR; (iii) DeepMinor (CNS2023-144375) and European Union
NextGenerationEU/PRTR. Olia Toporkov is supported by a doctoral grant from the UPV/EHU ``Formación de Personal Investigador'' (PIF 20/186). Rodrigo Agerri was additionally funded by the RYC-2017-23647 fellowship (MCIN/AEI/10.13039/501100011033 and by ESF Investing in your future).

\section{Bibliographical References}\label{sec:reference}

\bibliographystyle{lrec-coling2024-natbib}
\bibliography{anthology, bibliography}

\section{Language Resource References}
\label{lr:ref}
\bibliographystylelanguageresource{lrec-coling2024-natbib}
\bibliographylanguageresource{languageresource}

\onecolumn

\appendix

\textit{Appendix A. Detailed Out-of-Vocabulary Results} \label{sec:appendix}

\begin{table}[h]
\centering
\renewcommand{\footnotesize}{\fontsize{8pt}{10pt}\selectfont}
\begin{tabular}{l|l|r|r|r|c|r|c|r|c}
\hline
\multicolumn{4}{c|}{} & \multicolumn{2}{c|}{ses-udpipe} & \multicolumn{2}{c|}{ses-ixapipes} & \multicolumn{2}{c}{ses-morpheus} \\ \hline
& & \multicolumn{1}{c|}{oov} & \multicolumn{1}{c|}{oov} & \multicolumn{1}{c|}{oov} & \multicolumn{1}{c|}{oov lemmas} & \multicolumn{1}{c|}{oov} & \multicolumn{1}{c|}{oov lemmas} & \multicolumn{1}{c|}{oov} & \multicolumn{1}{c}{oov lemmas} \\ 
& & \multicolumn{1}{c|}{words} & \multicolumn{1}{c|}{lemmas} & \multicolumn{1}{c|}{ses} & \multicolumn{1}{c|}{(ses in train)} & \multicolumn{1}{c|}{ses} & \multicolumn{1}{c|}{(ses in train)} & \multicolumn{1}{c|}{ses} & \multicolumn{1}{c}{(ses in train)} \\ \hline
& ind & 7.85\% & 6.18\% & \textbf{0.02\%} & 99.89\% & 0.05\% & 99.74\% & 0.11\% & 99.07\% \\
\multirow{-2}{*}{es} & ood & 7.65\% & 5.93\% & \textbf{0.28\%} & 96.41\% & 0.43\% & 98.27\% & \textbf{0.28\%} & 97.86\% \\ \hline
& ind & 25.50\% & 13.74\% & \textbf{0.27\%} & 99.04\% & 0.67\%  & 98.31\% & 0.78\% & 97.35\% \\
\multirow{-2}{*}{ru} & ood & 29.21\% & 15.36\% & \textbf{1.65\%} & 96.23\% & 2.45\% & 95.03\% & 2.52\% & 94.33\% \\ \hline
& ind & 5.71\% & 4.19\% & \textbf{0.08\% }& 99.72\% & 0.10\% & 99.81\% & 0.25\% & 97.57\% \\ 
\multirow{-2}{*}{en} & ood & 11.89\% & 11.45\% & \textbf{1.32\%} & 90.41\% & 1.56\% & 91.58\% & 1.36\% & 91.15\% \\ \hline
& ind & 15.28\% & 5.07\% & \textbf{0.61\%} & 96.52\% & 1.45\% & 94.86\% & 0.92\% & 94.69\% \\
\multirow{-2}{*}{eu} & ood & 24.26\% & 11.99\% & \textbf{1.13\%} & 95.98\% & 2.49\% & 94.68\% & 1.45\% & 94.78\% \\ \hline
& ind & 24.83\% & 5.67\% & \textbf{0.12\% }& 99.69\% & 4.52\% & 95.69\% & 0.56\% & 97.23\% \\
\multirow{-2}{*}{tr} & ood & 36.71\% & 20.72\% & \textbf{0.45\%} & 97.85\% & 6.52\% & 91.40\% & 3.40\% & 97.85\% \\ \hline
& ind & 8.85\% & 3.19\% & \textbf{0.09\%} & 99.11\% & 0.20\% & 98.66\% & 0.24\% & 97.90\% \\
\multirow{-2}{*}{cz} & ood & 21.97\% & 11.76\% & \textbf{2.33\%} & 99.12\% & 2.59\% & 99.12\% & 2.90\% & 98.24\% \\ \hline
& ind & 19.53\% & 7.80\% & \textbf{0.28\%}& 99.22\% & 0.52\% & 98.82\% & 0.85\% & 97.84\% \\
\multirow{-2}{*}{pl} & ood & 17.65\% & 8.72\% & \textbf{0.40\%} & 98.65\% & 0.62\% & 97.57\% & 0.67\% & 97.57\% \\ \hline
                                          
\end{tabular}
    \caption{The proportion (in \%) of out-of-vocabulary words, lemmas and SES in the in-domain (ind) and out-of-domain (ood) test sets with respect to the train set, per language. In \textbf{bold}: lowest percentage of out-of-vocabulary (oov) SES among the three SES types.}
    \label{tab:oov_detailed}
\end{table}

Table \ref{tab:oov_detailed} reports the proportion of out-of-vocabulary (oov) words, lemmas and SES, both for in-domain (ind) and out-of-domain (ood) settings for the three SES types. By out-of-vocabulary we understand words, lemmas and SES in the test sets that the system did not see during the training process. The column \textit{`oov lemmas (ses in train)'} refers to the proportion of lemmas that the model does not see during the training (out-of-vocabulary lemmas) while their corresponding SES exist in the train set. In other words, they have been seen by the system.

\end{document}